\PassOptionsToPackage{main=english,russian}{babel}
\PassOptionsToPackage{T2A}{fontenc}
\documentclass[onecolumn]{misraj}

\usepackage{lipsum}
\usepackage{pdfpages}
\usepackage{colortbl}
\usepackage{tabulary}
\usepackage{longtable}
\usepackage{array}
\usepackage{placeins}
\usepackage{tablefootnote}
\usepackage{tcolorbox}
\usepackage{wrapfig}
\usepackage{adjustbox}
\usepackage{float}
\usepackage{caption}
\usepackage{breqn} 

\usepackage{pifont}
\definecolor{deeppurple}{HTML}{9e02f7}
\definecolor{forestgreen}{HTML}{2e7d43}
\usepackage{multirow}
\usepackage{tabularx}

\usepackage[utf8]{inputenc}

\usepackage{arydshln}

\usepackage{xspace} %
\usepackage{makecell} 
\usepackage{multirow}

\usepackage[framemethod=TikZ]{mdframed}
\usepackage{tcolorbox}
\usepackage{multicol}

\usepackage[utf8]{inputenc}
\usepackage[T1]{fontenc}
\usepackage[english]{babel} 
\usepackage{arabtex}
\usepackage{utf8}
\setcode{utf8}

\definecolor{lightblue}{RGB}{211, 227, 252} 
\definecolor{bgblue}{RGB}{247, 250, 255} 

\newcommand*\colourcheck[1]{%
  \expandafter\newcommand\csname #1check\endcsname{\textcolor{#1}{\ding{52}}}%
}
\newcommand*\colourcross[1]{%
  \expandafter\newcommand\csname #1cross\endcsname{\textcolor{#1}{\ding{55}}}%
}
\DeclareSymbolFont{extraup}{U}{zavm}{m}{n}
\DeclareMathSymbol{\vardiamond}{\mathalpha}{extraup}{87}

\colourcheck{blue}
\colourcheck{green}
\colourcheck{red}

\colourcross{blue}
\colourcross{green}
\colourcross{red}
\usepackage{nicematrix}
\usepackage{comment}
\usepackage{amssymb}
\usepackage{arabtex}
\usepackage[utf8]{inputenc}
\usepackage[bottom]{footmisc}
\RequirePackage[hidelinks,colorlinks=true,linkcolor=blue,citecolor=blue]{hyperref}
\setcitestyle{number,square}
\title{\raisebox{-0.2em}{\includegraphics[width=0.8cm]{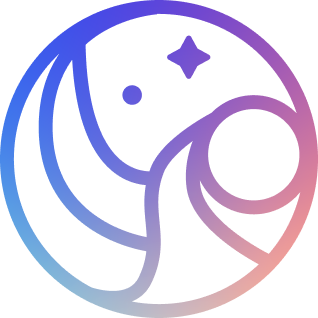}} Kuwain 1.5B: An Arabic SLM via Language Injection}

\multiauthors
\author{
    name={Khalil Hennara},
    email={hennara@misraj.ai}
}

\author{
    name={Sara Chrouf},
    email={sara.chrouf@misraj.ai}
}
\author{
    name={Mohamed Motasim Hamed \textsuperscript{*}},
    email={hamed@misraj.ai}
}
\author{
    name={Zeina Aldallal \textsuperscript{*}},
    email={aldallal@misraj.ai}
}
\author{
    name={Mohammad Omar Hadid \textsuperscript{\dag}},  
    email={m.omar.hadid.1997@gmail.com}
}
\author{
    name={Safwan AlModhayan},
    email={safwan@misraj.ai}
}

\affiliations{
    \includegraphics[width=2cm]{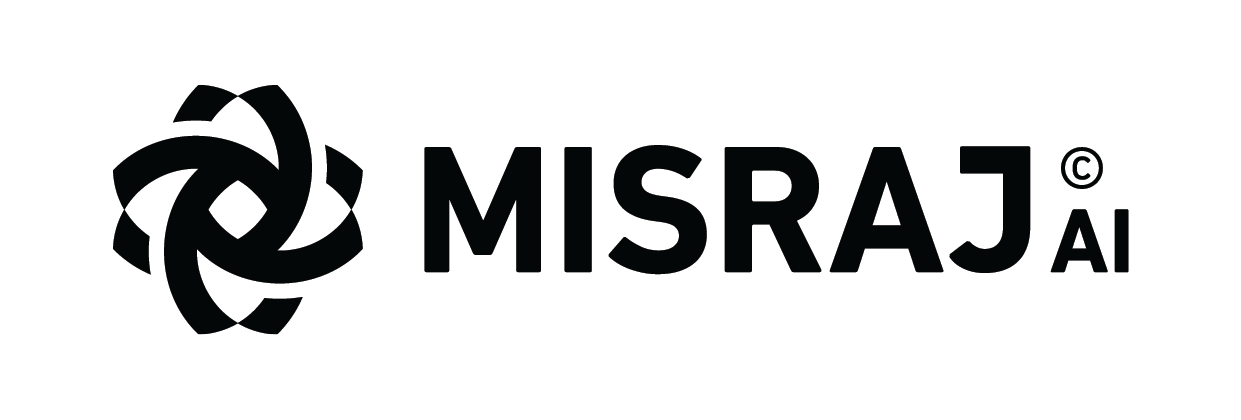}\\
     Khobar, Saudi Arabia \\
     {hennara,chrouf,hamed,aldallal,safwan}@misraj.ai \\
}
\date{\today}
\abstract{
Enhancing existing models with new knowledge is a crucial aspect of AI development. This paper introduces a novel method for integrating a new language into a large language model (LLM). Our approach successfully incorporates a previously unseen target language into an existing LLM without compromising its prior knowledge. We trained a tiny model with 1.5 billion parameters named \textit{\textbf{Kuwain}} \textsuperscript{\ddag} by injecting the Arabic language into a small open-source model mainly trained in English. Our method demonstrates significant improvements in Arabic language performance, with an average 8\% improvement across various benchmarks, while retaining the model's existing knowledge with a minimum amount of the original model's data. This offers a cost-effective alternative to training a comprehensive model in both English and Arabic. The results highlight the potential for efficient, targeted language model expansion without extensive retraining or resource-intensive processes.}

\begin{document}

\renewcommand{\thefootnote}{\fnsymbol{footnote}}
\footnotetext[1]{Equal contributions}
\footnotetext[2]{Formerly affiliated with misraj.ai at the time of this research; m.omar.hadid.1997@gmail.com }
\footnotetext[3]{\textbf{Kuwain: \<كُوَيْن>}: is a diminutive form of the Arabic word (Kawn), which means "universe". So, "Kuwain" means "tiny universe" or "little cosmos".Kuwain is one of a series of Arabic-English multilingual Large Language Models (LLMs), embodying the concept of a vast amount of knowledge condensed into a compact,  accessible form, much like a miniature universe of information.}

\section{Introduction}
\label{sec:introduction}
\begin{figure}[ht]
    \centering
    \includegraphics[width=\linewidth]{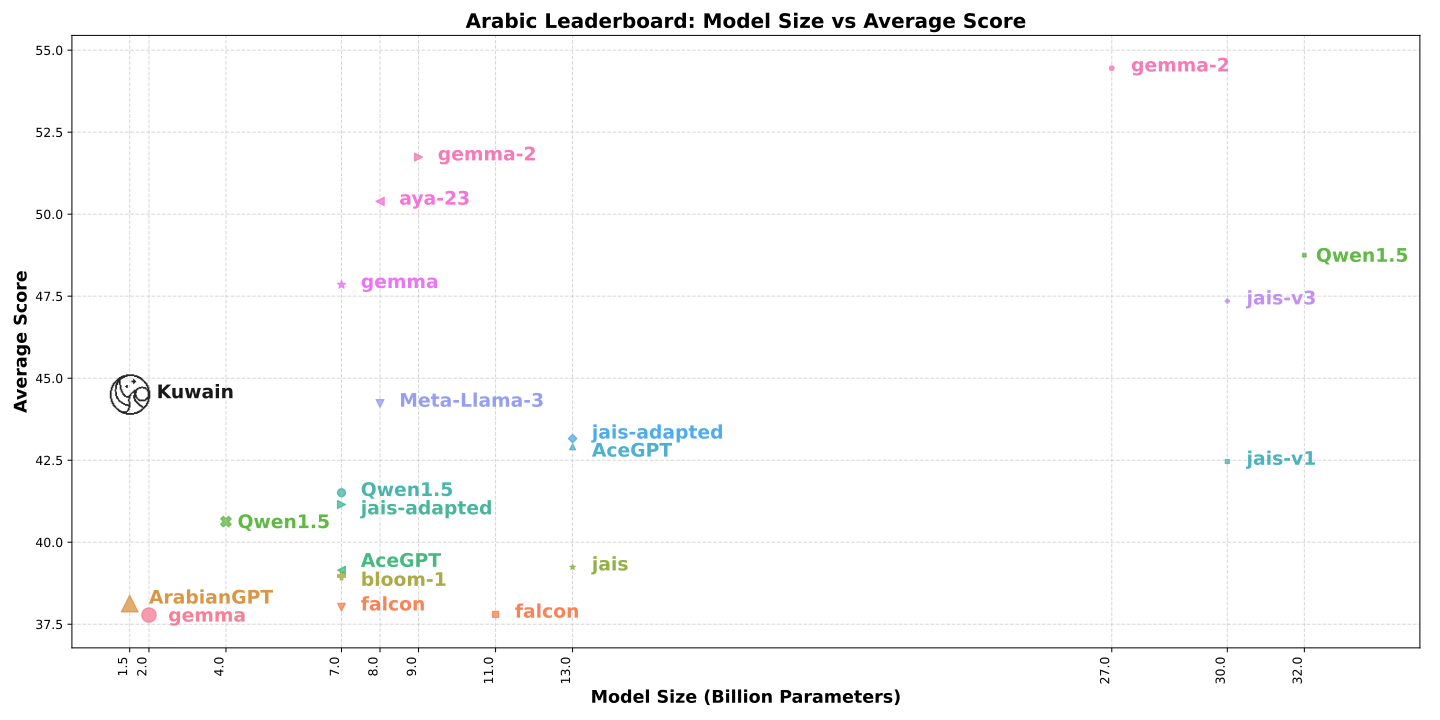}
    
    \caption{
        Model performance vs size visualization: Points represent models on the Arabic NLP benchmark, where point size indicates efficiency (average score divided by model size). Larger points show better performance-to-size ratio.
    }
    \label{fig:leader_bord}
\end{figure}




Large language models (LLMs) have achieved significant advances in recent years, demonstrating impressive performance across a wide range of natural language processing tasks \cite{touvron2023llama}. However, a notable limitation of these models is their English-centric nature \cite{zhang2023don}, with the majority focusing mainly on English tasks. This English bias makes language models (LMs) less effective in multilingual settings \cite{lai2023chatgpt}, particularly for languages with distinct linguistic characteristics and writing systems.

Despite the significant progress made by open-source models \cite{touvron2023llama,jiang2023mistral,abdin2024phi,team2024gemma,llama3} in natural language processing, they exhibit inherent limitations in supporting tasks in languages like Arabic. This may result in misunderstandings and impede the effectiveness of LMs when trained on Arabic data.

In this paper, we present \textit{\textbf{Kuwain 1.5B}}, a compact model with enhanced capabilities for understanding and generating Arabic language competing with much larger models, as illustrated in Figure \ref{fig:leader_bord}. To mitigate the high computational costs associated with training large language models, we propose a more efficient alternative which includes expanding a monolingual Small Language Model (SLM) with new Arabic language capabilities by training only additional layers on top of the open-source TinyLlama 1.1B model \cite{zhang2024tinyllama} instead of retraining all model's parameters. We incorporate a small amount of English data during training alongside our full Arabic dataset to preserve the model’s original English proficiency, eliminating the need for extensive retraining to maintain its original English capabilities. 
Previous research has demonstrated that using English-centric tokenizers in training multilingual LLMs can lead to significant downstream performance degradation and increase training costs by 68\% due to inefficient tokenization vocabularies \cite{ali2310tokenizer}. To address tokenization challenges, we expanded the vocabulary with additional $26K$ Arabic tokens. 
This research addresses two significant questions:
\begin{enumerate}
\item Expanding Language Support: How can we effectively and efficiently expand a monolingual LLM to support new languages while minimizing cost?
\item Preserving Original Language Performance: How can we maintain the original language performance and knowledge of an LLM while incorporating new language abilities, without compromising either?
\end{enumerate}

Our contributions can be summarized as follows:
\begin{itemize}
\item We propose a novel method for injecting a new language (Arabic) into an English-centric monolingual language model without retraining from scratch.
\item We present \textbf{Kuwain}, a compact multilingual Arabic-English language model designed to effectively process and understand a wide range of Arabic language tasks.
\end{itemize}

The results showed that our approach reduced training costs by 70\%, increased new language performance by 8\% compared to the original model, and maintained the original English language knowledge with a slight improvement of 1\%. Despite its relatively small size, \textit{Kuwain} has demonstrated strong performance in Arabic. A fine-tuned version of \textit{Kuwain}, called \textit{Lahajawi} \cite{hamed2025lahjawi}, has achieved impressive results in Arabic cross-dialect translation. We conclude from the above that our approach can be extended to different models of different sizes and with any new language without losing the inherent knowledge of the original model, and at a small cost with competitive performance.

The remainder of this paper is organized as follows: In Section~\ref{sec:related}, we review the related work, providing an overview of multilingual large language models (MLLMs) and exploring previous efforts to develop Arabic-centric LLMs. The description of our training data is in section~\ref{sec:data}. Section~\ref{sec:method} details our proposed method, including the extension of model layers and vocabulary expansion. In Section~\ref{sec:exp}, we detail our experimental setup and results. Section~\ref{sec:evaluation} offers a comprehensive evaluation of our model's performance compared to existing open-source models. Finally, in Section~\ref{sec:conclusion}, we summarize our findings and suggest directions for future research.

\section{Related work}
\label{sec:related}

This section explores the methodologies for acquiring new languages in LLMs and provides an overview of recent advancements in Arabic LLMs. We begin by discussing various approaches to pre-training and extending LLMs to handle multiple languages, focusing on strategies like continuous pre-training and vocabulary expansion. Following this, we present an overview of key Arabic LLMs, examining their approaches to improve linguistic accuracy, cultural relevance, and performance in Arabic natural language processing.

\subsection{Multilingual LLMs Overview}

\subsection*{ \textit{Pre-training LLMs From-Scratch}}

Several studies \cite{touvron2023llama,jiang2023mistral,chowdhery2023palm, le2023bloom, xue2020mt5} incorporated multilingual data during the pre-training phase to improve model alignment. As \cite{blevins2022language} observed, even the unintentional inclusion of a small amount of multilingual data during initial pre-training can significantly enhance multilingual performance.
Training LLMs from scratch presents significant challenges due to the complexity of optimizing randomly initialized parameters \cite{touvron2023llama, llama3}. This optimization process demands extensive training data to guide the model toward effective performance \cite{kaplan2020scaling, brown2020language}.

\subsection*{\textit{Language Extension in LLMs}}

Multilingual large language models can process and generate content in multiple languages simultaneously, such as English and Arabic. Integrating new languages into existing LLMs is a promising area of exploration \cite{shaham2024multilingual}.
Continuous pre-training of LLMs involves extending the pre-training process on already pre-trained models to reduce computational costs and improve efficiency. This approach is critical for keeping LLMs up-to-date with new information and specialized domains without the need for full retraining. It allows models to remain relevant and effective in various applications while being more computationally efficient \cite{gupta2023continual}. For instance, 
\cite{cui2023efficient} suggests adding languages through a two-stage pre-training process, emphasizing the extension of the LLMs’ vocabularies to accommodate new languages.
Similarly,  \cite{pluster2023leolm} released LeoLM-7B, a German foundation model developed through continual pre-training on Llama2-7B. Comparable work has also been done for Japanese \cite{okazaki2024building}. Both models were created through continual pre-training on Mistral-7B and Meltemi-7B, the first open LLM for the Greek language \cite{voukoutis2024meltemi}. This work adapts the Mistral model to new languages using a continuous pre-training approach. Their method includes re-warming and re-decaying the learning rate while incorporating Greek and English monolingual data into the pre-training dataset. Results show that models subjected to continuous pre-training tend to perform worse than their base models on original language tasks. For example, Meltemi-7B performs worse than Mistral-7B on English tasks by -6\% \cite{voukoutis2024meltemi}. \cite{pluster2023leolm} reports that LeoLM-7B achieved an average improvement of +4.8\% for German benchmarks compared to the base Llama-2 model, while it scored lower by -2.6\% on English benchmarks. Swallow-MS-7b-v0.1 \cite{okazaki2024building} showed an average improvement of +8\% on Japanese benchmarks versus its base model but had lower average scores of -5.3\% on English benchmarks.
Furthermore, \cite{blevins2024breaking} employed a novel approach by leveraging the Mixture-of-Experts (MoE) technique. This method involves training separate language models on distinct subsets of multilingual corpora, allowing for independent model training and effectively mitigating issues arising from competition between multiple languages within a single model's parameters. Additionally, \cite{zhou2024moe} designed MoE-LPR, a Mixture-of-Experts approach with Language Priors Routing, to enhance the multilingual capabilities of large language models. This approach mitigates the issue of catastrophic forgetting by preserving original language knowledge and improving performance on expanded languages. \cite{yong2022bloom+} observes that Adapter-based methods are more effective than continuous pre-training.

\subsection{Arabic-centric LLMs Overview}

The field of Arabic Large Language Models (LLMs) has made notable progress recently, with several models developed to address the specific challenges and opportunities of the Arabic language. This overview focuses on key models: Jais \cite{sengupta2023jais}, AceGPT \cite{huang2023acegpt}, ArabianGPT \cite{koubaa2024arabiangpt} and ALLaM \cite{saiful2024allam}. Each model takes a unique approach to Arabic natural language processing, aiming to balance linguistic accuracy, cultural relevance, and technical capabilities. The comparison of these models highlights the advancements and the ongoing challenges in creating effective Arabic LLMs, offering insights into the current state of Arabic language AI and the different strategies being pursued to improve its performance and cultural alignment.

In the evolving landscape of natural language processing, Jais \cite{sengupta2023jais} is an open-source, bilingual model trained from scratch to handle multiple tasks in Arabic and English. Its capabilities are further enhanced by Jais-chat, an instruction-tuned variant optimized for task-oriented dialogues. However, a significant limitation of Jais lies in its training data composition: the model has been exposed to substantially more English data than Arabic, with a large portion of the Arabic data being translated from English sources. This imbalance not only incurs high computational and financial costs but also leads to outputs that are often biased toward English cultural contexts, frequently neglecting the richness of Arabic’s artistic and literary heritage.

Recognizing this gap, AceGPT \cite{huang2023acegpt} and ArabianGPT \cite{koubaa2024arabiangpt} have taken steps to align these models with Arabic values and cultural contexts. AceGPT, in particular, employs a strategic approach by adapting the English-focused LLaMA2 model through additional pre-training, instruction tuning, and Reinforcement Learning from AI Feedback (RLAIF) without vocabulary expansion. However, the absence of vocabulary extension poses a notable limitation, as the original LLaMA2 tokenizer, being character-level for Arabic, lacks adequate coverage of Arabic morphology and vocabulary, leading to inefficiencies in token representation and model performance. Furthermore, like many bilingual models, AceGPT has been trained on significantly more English data than Arabic, which can introduce biases and limit its effectiveness in fully capturing the nuances of the Arabic language and culture. These challenges, despite the carefully designed interventions, highlight the need for more balanced and linguistically aware modeling strategies.

ArabianGPT \cite{koubaa2024arabiangpt}, based on the GPT-2-small architecture, is notable for its advanced tokenizer, AraNizer, which is optimized for Arabic's linguistic features, offering accurate text segmentation. However, its exclusive focus on Arabic presents limitations compared to bilingual models like AceGPT  \cite{huang2023acegpt} or ALLaM \cite{saiful2024allam}, which benefit from knowledge transfer from English. ArabianGPT may struggle with multilingual tasks, dialectal variations, and complex language tasks due to its bias toward Modern Standard Arabic, smaller model size, and limited training data.

ALLaM model series \cite{saiful2024allam} is designed to enhance Arabic language processing by leveraging a mixture of Arabic and English data. It incorporates vocabulary expansion and continued pre-training on 4 trillion English tokens and 540 billion Arabic tokens, achieving state-of-the-art performance in various Arabic benchmarks while maintaining proficiency in English. 

While existing efforts in multilingual and Arabic-centric LLMs have demonstrated substantial progress, they commonly suffer from three recurring limitations: (1) reliance on extensive resources and high training costs, (2) degradation in original language performance due to catastrophic forgetting, and (3) inadequate tokenization strategies, especially for morphologically rich languages like Arabic. Many models either fail to balance bilingual performance or demand substantial retraining, which poses challenges for scalability and accessibility. 
Our work introduces a lightweight and cost-efficient alternative. Unlike prior models that require full retraining or compromise original language capabilities, \textit{Kuwain} extends an existing English-centric LLM with minimal data and computational resources while preserving its original performance. By strategically expanding the tokenizer and adapting only specific model layers through continual pre-training, \textit{Kuwain} achieves significant performance gains in Arabic without the common trade-offs seen in prior approaches. This positions our method as a promising solution for multilingual expansion, particularly for underrepresented languages in resource-constrained environments. 

\section{Data}
\label{sec:data}
\subsection{Data Sources}
Our training dataset comprises 110 billion tokens, with 90 billion tokens in Arabic and 20 billion in English, all derived from publicly available open-source sources: CulturaX \cite{culturax}, C4 \cite{c4}, and ArabicText 2022\footnote[5]{\textbf{ArabicText 2022:} an open-source Arabic collection prepared by the Beijing Academy of Artificial Intelligence (BAAI), which includes Arabic corpora such as ArabicWeb22-A, ArabicWeb16, OSCAR, ArabicWeb22-B, CC100-AR, and Arabic Tweets.}. In addition to Modern Standard Arabic, the dataset also includes a portion of Arabic dialectal data sourced from Hugging Face repositories to preserve the linguistic richness and diversity of regional Arabic varieties. Importantly, none of the evaluation datasets were included in the training data, preventing data leakage and ensuring the integrity of our evaluation.

\subsection{Data Cleaning}
To improve the overall quality of the Arabic data, we applied a range of filtering and cleaning steps. These included removing corrupted or unreadable characters and repeated characters, stripping markup and elongation characters, and white space normalization, while retaining non-Arabic characters that may appear within the text. Additionally, we preserved Quranic symbols and other special characters commonly found in Arabic texts to maintain their integrity.
In addition, we normalized encoding inconsistencies and unified orthographic variants (e.g., different forms of \textit{baa}), Malformed examples and short text were also filtered out to ensure consistency and improve data quality.

This cleaning process was implemented using a custom script, which we released to support reproducibility and future improvements in Arabic text pre-processing. The script includes both the specific cleaning steps used in this study and general-purpose Arabic text cleaning functions, which are fully configurable, making it suitable for broader Arabic text pre-processing applications\footnote[4]{\url{https://github.com/misraj-ai/Kuwain-Arabic-cleaner}}. For English data, we adopted the filtering pipeline introduced by the BLOOM project \cite{le2023bloom}, which promotes clean, diverse, and high-quality content.

\section{Method}
\label{sec:method}
Our approach integrates a new language into an existing LLM while fully preserving its prior knowledge. This approach doesn't maintain the model's existing capabilities only, but also enriches and expands them. It is based on two key concepts: \textbf{(1)} extending the model architecture and \textbf{(2)} expanding the tokenizer's vocabulary. 
\begin{figure}[ht]
    \centering
    \begin{minipage}[t]{0.48\textwidth}
        \centering
        \includegraphics[width=\linewidth]{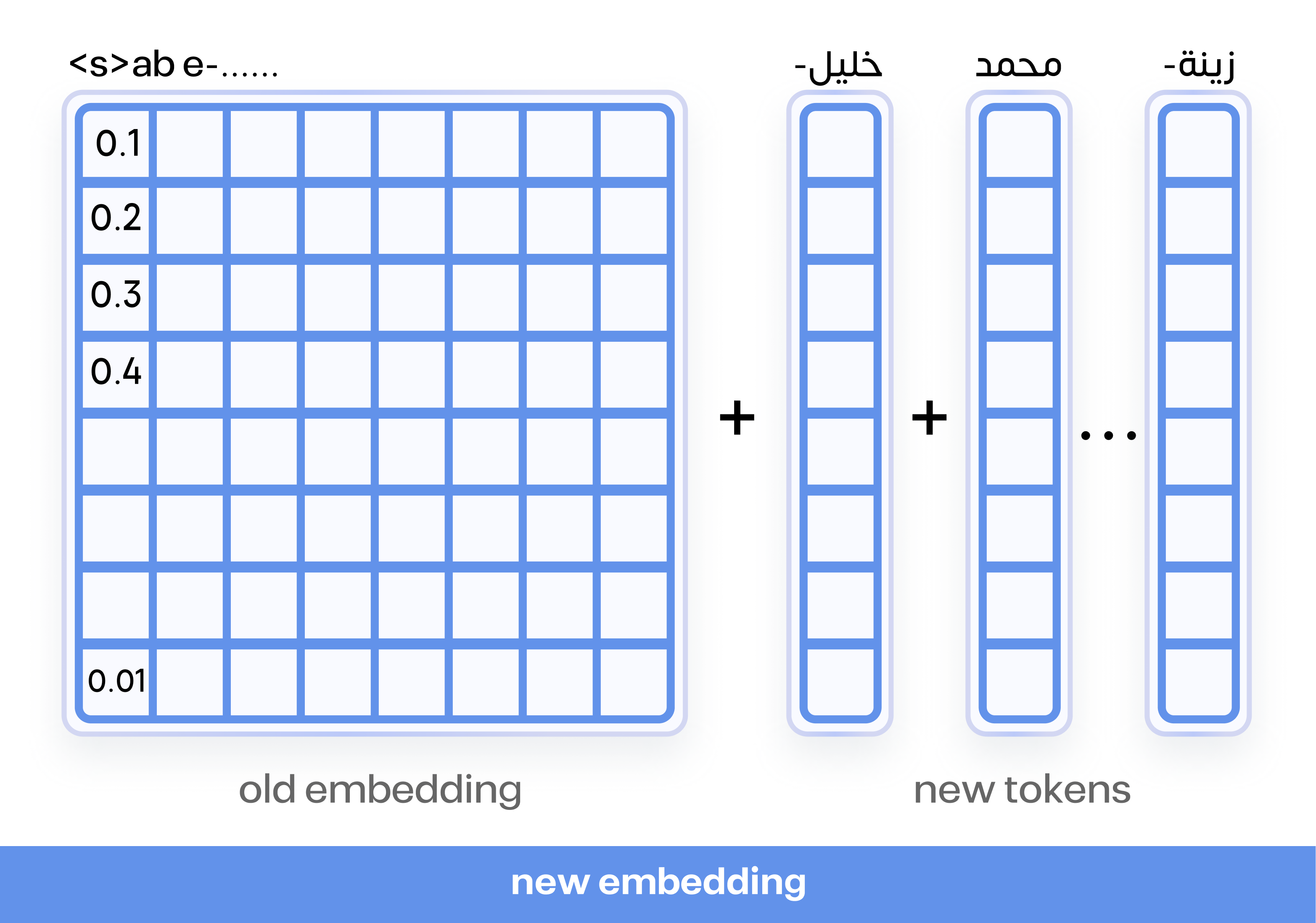}
        \caption*{(a)}
    \end{minipage}
    \hfill
    \begin{minipage}[t]{0.48\textwidth}
        \centering
        \includegraphics[width=\linewidth]{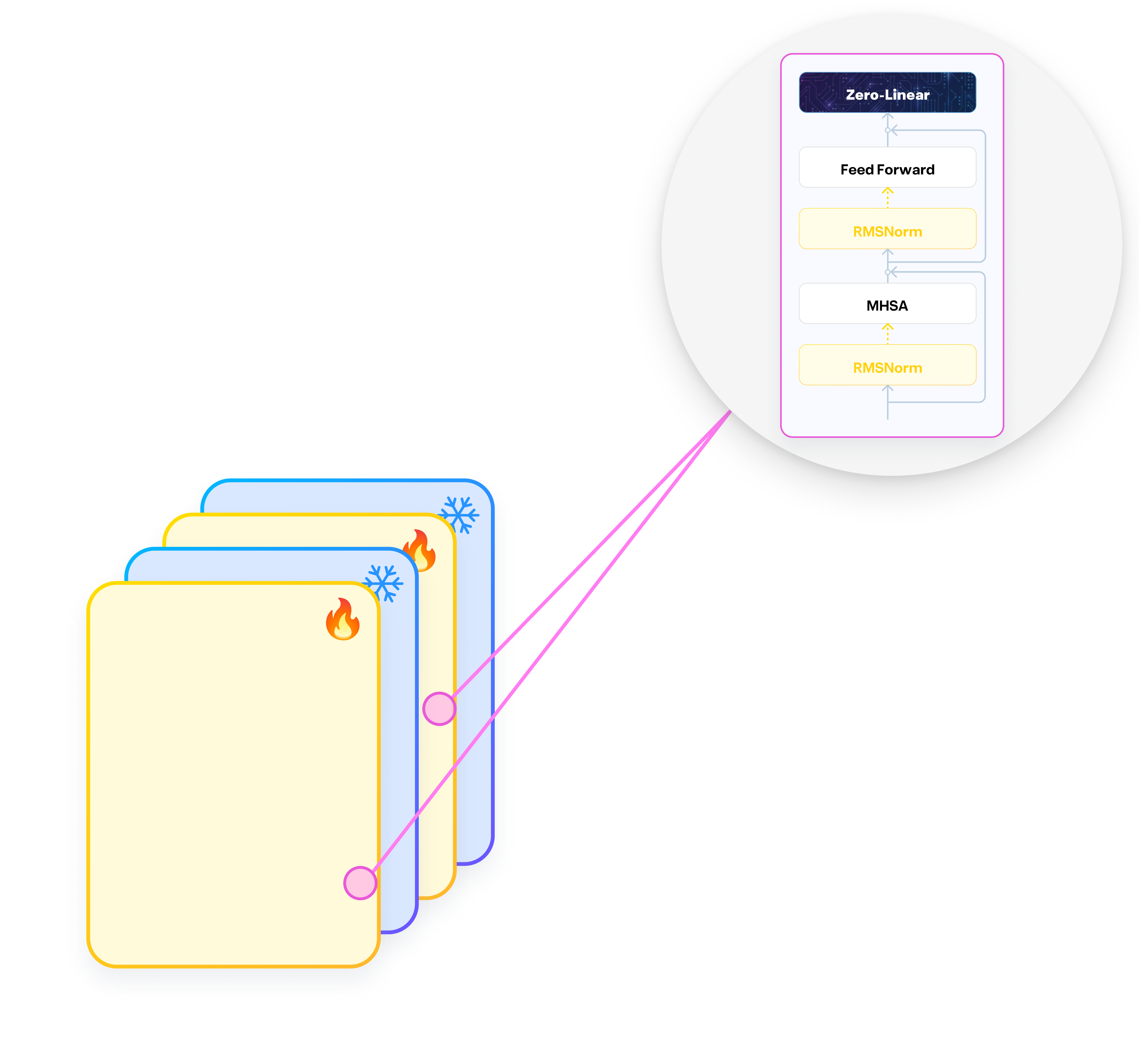}
        \caption*{(b)}
    \end{minipage}
    
    \caption{(a) Vocabulary expansion by adding new Arabic tokens to the tokenizer. (b) Extension of model layers, where newly added layers were trainable while the original layers remained frozen.}
    \label{fig:method}
\end{figure}

\subsection{Extending Model's Layers}
\begin{figure}[ht]
\centering
\includegraphics[width=0.7\textwidth]{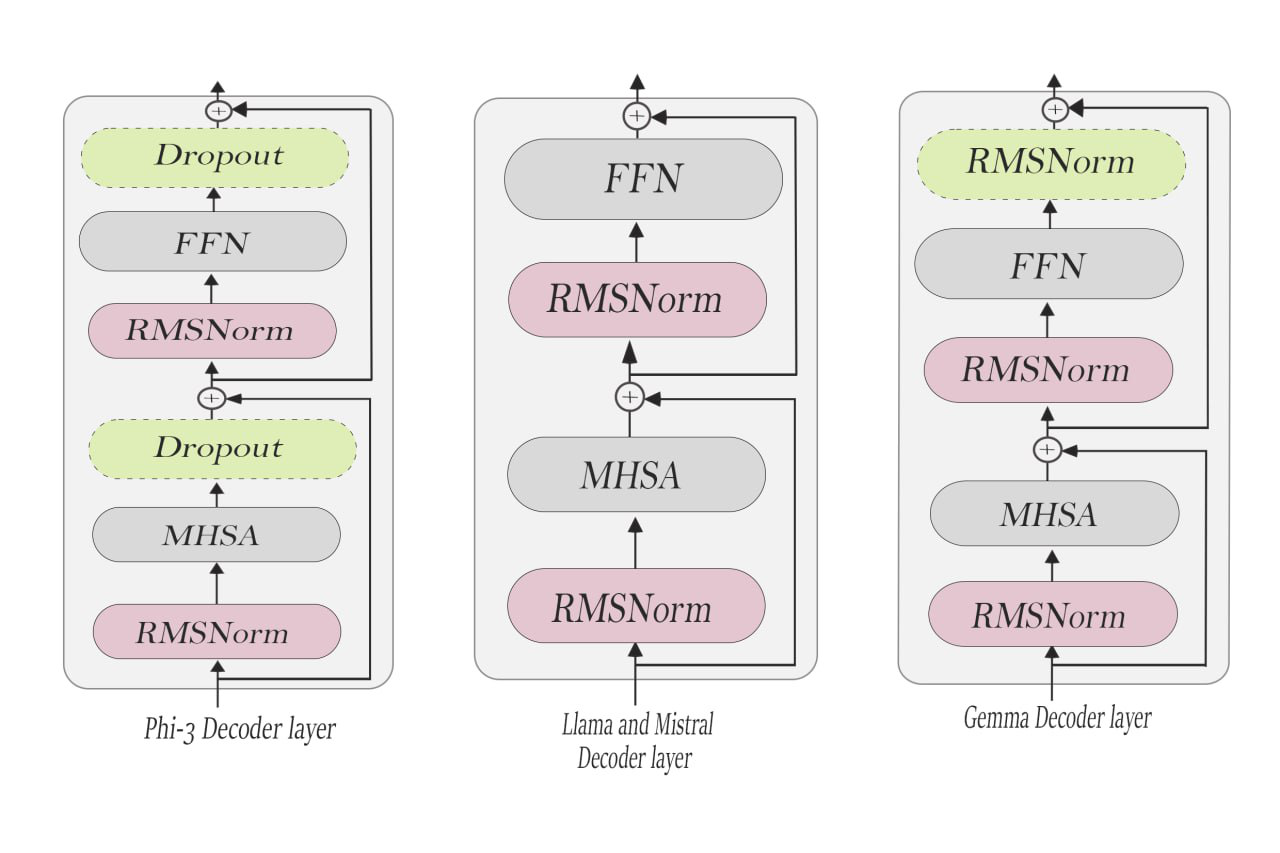}
\caption{models decoder architecture.}
\label{fig:decoders_arc}
\end{figure}

Inspired by \textit{Llama-Pro} \cite{wu2024llama}, which reaches state-of-the-art performance across a broad range of general, code, and math tasks, they explored extending the model's capabilities by injecting a new layer every n layers. We adopt a similar approach but with a broader experimental scope. The Llama-Pro approach introduces the concept of layer extension, which incorporates an identity layer after a group of existing layers in the original model. This technique can be applied to various open-source models, such as Llama-2,3  \cite{touvron2023llama, llama3}, Mistral \cite{jiang2023mistral}, Phi-3 \cite{abdin2024phi}, and Gemma-2 \cite{team2024gemma}, which share similar decoder layer structures with minor variations. These minor modifications do not significantly alter the identity block concept as described in \cite{wu2024llama}. Figure \ref{fig:decoders_arc} illustrates the common structure of these decoder layers.

A typical decoder block in these models produces an output $y$ from an input $x$ as defined by the following equations:
$$x' = x + MHSA(NormLayer(x))$$
$$x'' = x' + FFN(NormLayer(x'))$$

Here, \textbf{MHSA} stands for Multi-Head Self-Attention, \textbf{FFN} is the Feed-Forward Network, and \textbf{NormLayer} is a normalization layer (e.g., LayerNorm or RMSNorm). The internal structures of \textbf{MHSA} and \textbf{FFN} are defined as:

\begin{equation} \label{equ:one} \text{FFN}(x) = \text{Activation}(x, W^1) W^2 W^3 \end{equation}
\begin{equation} \label{equ:two} \text{MHSA}(Q, K, V) = \text{Concat}(\text{head}_1, ..., \text{head}_h) W^O \end{equation}

Where $W^1, W^2,W^3$ are the learnable weight matrices in the \textbf{FFN}, and the activation function may vary depending on the model. For instance, LLaMA \cite{touvron2023llama} uses SwiGLU \cite{shazeer2020glu}, whereas Gemma-2 \cite{team2024gemma} uses a Linear activation. In the attention mechanism, $W^O$ is the output projection matrix.

To extend the model architecture, we build upon this standard decoder block by introducing additional blocks while preserving the original model’s behavior. Suppose the model consists of blocks $(\theta_{0},\theta_{1},...,\theta_{n} )$. The extended model includes additional identity blocks $\theta_{id}$, defined as:
$$\theta_{id}(x) = x$$
This ensures the input and output of the identity block are the same. The identity behavior can be achieved if the following conditions are satisfied:
\begin{equation}
    MHSA(NormLayer(x))=0, FFN(NormLayer(x^{\prime}))=0
\end{equation} 

These conditions can be met by initializing the additional blocks' weight matrices $W^O$ and $W^3$ in Equations \ref{equ:one} and \ref{equ:two} to zero. This ensures that the new blocks initially perform no transformation on the input.
It's important to note that certain model-specific variations exist:
    Gemma-2 applies an additional RMSNorm after the MLP:
    $$y=LayerNorm(x'')$$

    Phi-3 includes a Dropout layer after the MLP:
    $$y=x'+Dropout(x'')$$

These minor modifications do not affect the identity property described above.

While \cite{wu2024llama} focuses on enhancing model performance on tasks it has previously encountered, our work addresses a more challenging objective: adapting a model to the Arabic language, which is absent from its original pretraining data.  Adapting a model to Arabic, a language absent from its original pre-training data. Whereas the extended layers in \textit{Llama-Pro} are specialized for domain-specific knowledge, the extended layers in our work are tailored to the injected language. In addition to this language adaptation, we extend the model's vocabulary by introducing several new tokens. Our full methodology is illustrated in Figure \ref{fig:method}.

\subsection{Vocabulary Expansion}

Extending a tokenizer’s vocabulary is a critical step when adapting a model to a new language particularly when the original tokenizer, such as a character-based one, is not well-suited for that language \cite{cui2023efficient, voukoutis2024meltemi, csaki2023efficiently, nikolich2024vikhr}. Inspired by the above we conclude that vocabulary expansion improves tokenization efficiency, enabling the model to represent the language more effectively. This not only enhances performance but also reduces training costs by shortening the number of required training steps and increasing the effective context length.

The first step in extending a tokenizer's vocabulary is to analyze the new language's text data to assess how well the existing tokenizer covers it. This involves identifying gaps where the current vocabulary fails to capture important characters, words, or subwords. Many open-source models, such as \textbf{LLaMA-2} \cite{touvron2023llama}, \textbf{Phi-3} \cite{abdin2024phi}, and \textbf{Mistral} \cite{jiang2023mistral}, are primarily designed for English and include only 28 Arabic tokens—covering just the Arabic alphabet. This limited representation is inadequate for capturing the morphological richness and lexical diversity of Arabic. In contrast, the \textbf{Llama-3} \cite{llama3} and \textbf{Gemma} \cite{team2024gemma} includes over 4,000 Arabic tokens, offering better coverage. However, even this may not fully reflect the complexity and nuance of Arabic \cite{farghaly2009arabic, elkateb2006arabic, issa2023morphological}.

In our work, we have trained a tokenizer using \textbf{SentencePiece} \cite{kudo1808sentencepiece} with our training data, resulting in a vocabulary of \textbf{26K} tokens. The vocabulary from this tokenizer is then used to extend the vocabulary of the base model's tokenizer, resulting in 54K tokens. We conducted an extensive evaluation to ensure that these 26K tokens are optimized for our model; see section~\ref{sec:tokenizer_exp} for more details. 

\section{Experiments and Results}
\label{sec:exp}
We conducted experiments using \textit{TinyLlama} \cite{zhang2024tinyllama} as our base model to evaluate Arabic language integration capabilities. \textit{TinyLlama}, a compact 1.1B language model pretrained on around 1 trillion tokens, demonstrating remarkable performance despite its size. The experimental architecture involved adding 8 new layers and extending the vocabulary with 26K Arabic tokens while keeping the original layers frozen. Training was performed on 8 A100 GPUs for 3 epochs, and the effective batch size is 1M tokens, focusing only on the newly added layers to preserve existing model knowledge while developing Arabic language capabilities. 

\subsection{Number and Placement of New Layers}
We conducted comprehensive experiments to evaluate both the number and positioning of the inserted layers, exploring configurations ranging from 6 to 10 layers. A subset of our dataset was used for this investigation, with all experiments related to layer count and placement performed over 20K training steps. Additionally, we assessed the model’s language generation capabilities through a human evaluation of a 500-sentence completion task . Unlike the \textit{LLaMa Pro} approach, which added layers after equal-sized blocks, we found no significant advantage to this uniform distribution method. Our findings reveal several key insights:

\begin{figure}[ht]
    \centering
    \begin{minipage}[t]{0.32\textwidth}
        \centering
        \includegraphics[width=\linewidth]{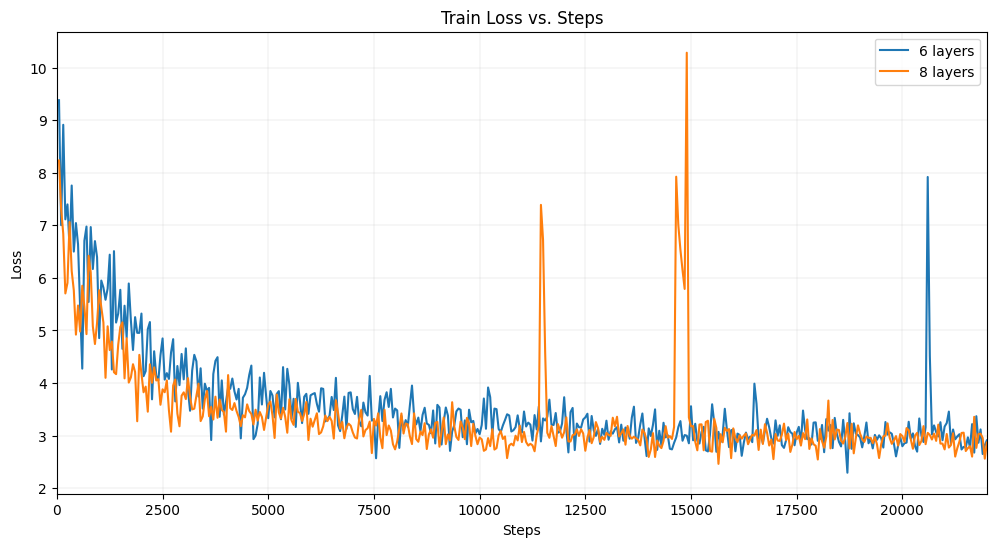}
        \caption*{(a)}
    \end{minipage}
    \hfill
    \begin{minipage}[t]{0.32\textwidth}
        \centering
        \includegraphics[width=\linewidth]{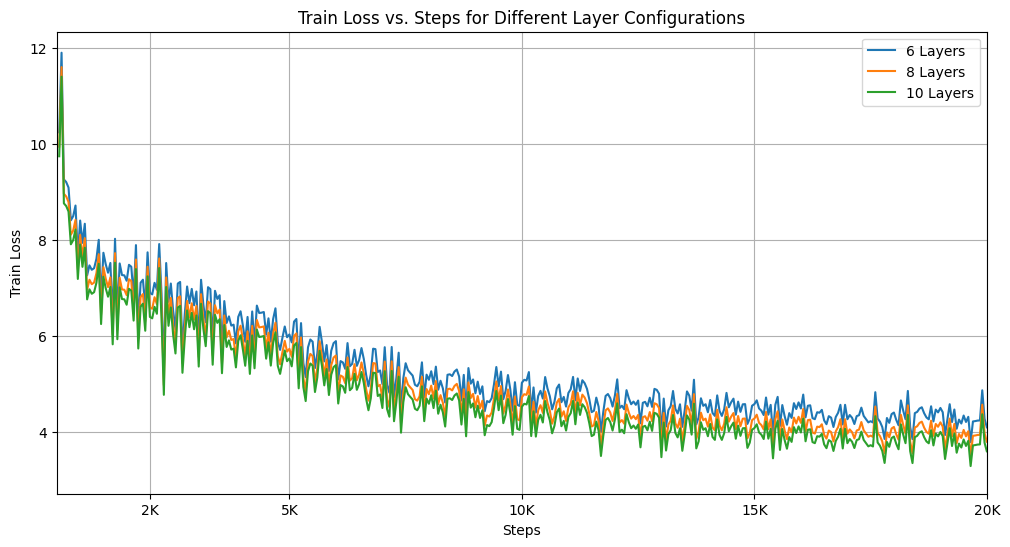}
        \caption*{(b)}
    \end{minipage}
    \hfill
    \begin{minipage}[t]{0.32\textwidth}
        \centering
        \includegraphics[width=\linewidth]{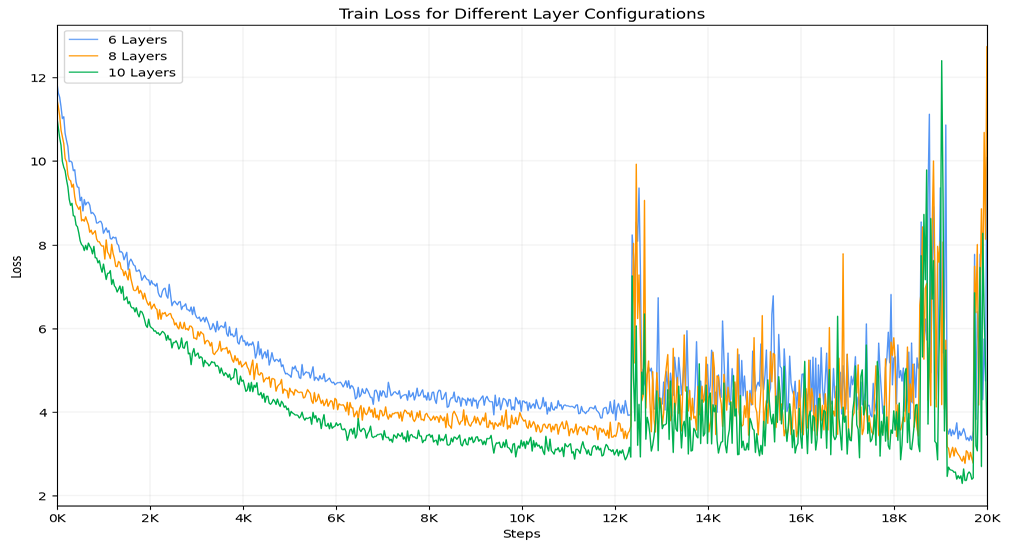}
        \caption*{(c)}
    \end{minipage}
    
    \caption{
        Training loss charts for Kuwain under different configurations: 
        (a) Freezing the last layer leads to instability, 
        (b) Enabling the last layer and avoiding consecutive layers yields stable training, 
        (c) Stacking new layers consecutively results in degraded performance.
    }
    \label{fig:combined_charts}
\end{figure}

\begin{enumerate}
    \item Consecutive insertion of new layers should be avoided, as it can lead to training instability.
    \item The final encoder layer must remain trainable to ensure stable training outcomes. This aligns with the findings of \cite{gromov2024unreasonable}, which emphasize the importance of maintaining the final layer across all pruning sizes. In our case, however, we discovered that the last layer specifically needs to be trainable to prevent volatility in the training process.
    \item The optimal placement of new layers does not necessarily follow a regular pattern or equal intervals between existing blocks.
    \item We observed no significant differences in performance between models with 8 to 10 new layers. However, the best results were achieved using 8 new layers, which represents an increase of approximately 30\% in model size compared to the original architecture.
\end{enumerate}

Figure~\ref{fig:combined_charts} illustrates findings from several training configurations for \textit{kuwain}. The experiments cover the following scenarios:

\begin{itemize}
    \item \textbf{Trainable vs. frozen last layer:} Comparing the effect of allowing the final layer to update during training versus keeping it fixed.
    \item \textbf{Stacking multiple new layers:} Assessing performance and stability when several new layers are added consecutively.
    \item \textbf{Distributed insertion of new layers:} Testing a configuration where 8 new layers are spread across the model, rather than stacked.
\end{itemize}

As shown in chart~\textit{(a)}, freezing the last layer led to unstable training, highlighting the importance of keeping it trainable. In \textit{(c)}, stacking multiple new layers significantly worsened performance, emphasizing the drawbacks of this strategy. The charts collectively demonstrate how these architectural choices affect training stability and performance, with the distributed 8-layer configuration emerging as the most effective.

\subsection{Tokenizer Comparison}
\label{sec:tokenizer_exp}
In this analysis, we evaluate our proposed tokenizer against two prominent Arabic language models: \textit{AraBERT} \cite{antoun2020arabert} and \textit{Jais} \cite{sengupta2023jais}. This comparative study centers on two critical metrics that impact the performance and efficiency of a tokenizer:

\begin{enumerate}
    \item \textbf{Vocabulary Size:} This refers to the number of unique tokens in the tokenizer's vocabulary. It is a pivotal metric, as it determines the tokenizer's ability to capture and represent diverse linguistic features within a language. A larger vocabulary may provide more granularity but also introduces the risk of overfitting and inefficiency in model training.
    
    \item \textbf{Expansion Ratio:} Defined as $\frac{L_{\text{tokenized}}}{L_{\text{original}}}$, where $L_{\text{original}}$ is the length of the original text measured in words and $L_{\text{tokenized}}$ represents the number of tokens after tokenization. This metric indicates the degree of increase in text length due to the tokenization process. An expansion ratio greater than 1 suggests that tokenization results in a longer text than the original.
\end{enumerate}

An ideal tokenizer balances a reasonable expansion ratio with a manageable vocabulary size. While a large vocabulary can enhance linguistic precision, it may also lead to computational inefficiency, poor generalization, and sparsity, particularly in morphologically rich languages like Arabic. Therefore, vocabulary size must be carefully optimized to maintain both effectiveness and efficiency.

Building on our earlier evaluation of tokenizer performance, our findings align with the recent comprehensive study by \cite{tao2024scaling}, which offers an in-depth analysis of tokenizers' characteristics in the context of Large Language Models (LLMs). This alignment further emphasizes the importance of a well-designed tokenizer in the broader scope of NLP applications.

In line with the conclusions drawn by \cite{tao2024scaling}, we trained multiple tokenizers and fine-tuned their parameters to identify the optimal one for the vocabulary expansion task. Our primary objective is to minimize the number of tokens while ensuring that the core meaning of the text is preserved. To evaluate and compare the performance of these tokenizers, we adopt the expansion ratio metric, as recommended by \cite{bostrom2020byte}, to determine which tokenizer achieves the best balance. Table \ref{tab:tokenizer_comparison} presents the results of the tokenizers we trained, alongside those of the \textit{Jais} \cite{sengupta2023jais} and \textit{AraBERT} \cite{antoun2020arabert} tokenizers, providing a comparison of their performance and the size of their Arabic vocabularies.
\begin{table}[htbp]
\centering
\fontfamily{ptm}\selectfont
\fontsize{10}{12}\selectfont
\begin{tabular}{lrrr}
\toprule
\textbf{Model}  & \textbf{Expansion Ratio} & \textbf{Vocabulary Size} \\
\midrule
\textit{Kuwain Tokenizer} &  2.30 & 26K \\
\hline
AraBERT &  2.51 & 54K \\
Jais &  2.19 & 44K \\
\midrule
\end{tabular}
\caption{ Comparison of tokenization metrics across different Arabic language models. The vocab size used presents the \textit{Arabic} vocab for each model, where the full vocab has a different number.}
\label{tab:tokenizer_comparison}
\end{table}

This analysis is based on the tokenization of over 1 million examples from our datasets. The results reveal that while Jais \cite{sengupta2023jais} achieves the best expansion ratios, our tokenizer demonstrates superior efficiency when considering the trade-off between this metric and vocabulary size. This balance suggests that our tokenizer achieves effective text representation without the need for an expansive vocabulary, potentially leading to more efficient model training and deployment.

\subsection{Data Proportion}
Our experiments highlight the importance of properly mixing old and new data when adapting language models to new languages. We discovered that Kuwain requires only 20\% of the original English data to maintain performance on English tasks while successfully acquiring Arabic language capabilities. This is significantly less than the 50\% ratio used in comparable approaches \cite{cui2023efficient, voukoutis2024meltemi, saiful2024allam, csaki2023efficiently}.

This smaller proportion of familiar data serves as an "anchor" that helps the model retain its existing knowledge while efficiently learning new skills.

To assess the impact of reduced exposure to the original language, we trained the model using the full Arabic dataset while limiting English data to less than 20\%. Table \ref{tab:merged_english_data_evaluation} shows that reducing the proportion of English data below 20\% (as in \textit{Kuwain-lt-$\phi$}) results in a notable decline in performance on English benchmarks. For example, the average score drops from 52.99 to 49.56. In contrast, training \textit{Kuwain} with 20\% English data is sufficient to maintain performance comparable to \textit{TinyLlama}—and in some cases, slightly better. This suggests that a minimal amount of original language data is essential to retain previously acquired capabilities.

These findings underscore the delicate balance in multilingual model adaptation: too little original-language data leads to knowledge degradation, while too much can hinder the learning of the new language and increase training costs.

\begin{table}[H]
\centering
\fontsize{10}{12}\selectfont
\captionsetup{justification=centering}
\begin{tabular}{lccc}
\hline
\textbf{Dataset} & \textbf{TinyLlama} & \textbf{Kuwain} & \textbf{Kuwain-lt-$\phi$} \\
\hline
HellaSwag     & \textbf{59.20} & 57.79 & 55.79 \\
Obqa          & \textbf{36.00} & 35.60 & 32.60 \\
ARC-c         & 30.10 & \textbf{30.29} & 27.29 \\
ARC-e         & 55.25 & \textbf{56.31} & 51.31 \\
boolq         & 57.83 & \textbf{60.43} & 54.43 \\
piqa          & \textbf{73.29} & 72.63 & 69.63 \\
WinoGrande    & 59.12 & \textbf{59.99} & 55.99 \\
\textbf{Avg}  & 52.99 & \textbf{53.28} & 49.56 \\
\hline
\end{tabular}
\caption{Performance comparison on \textbf{English} benchmarks between TinyLlama and two variants of Kuwain: trained with 20\% English data (Kuwain) and with less than 20\% English data (Kuwain-lt-$\phi$). Results show how English performance varies with data proportion.}
\label{tab:merged_english_data_evaluation}
\end{table}

\subsection{Conventional Training (Without Layer Extension)}
To assess the impact of conventional approaches in the absence of our proposed strategy—namely, language extension through continued pre-training and vocabulary expansion without layer extension—we trained the base model on the complete Arabic dataset, extending its vocabulary using the same tokenizer but without applying any layer extension. We refer to this baseline model as \textbf{Kuwain-Naive.} 

While \textit{Kuwain-Naive} demonstrated effective acquisition of new linguistic information, particularly in Arabic, it suffered a significant loss in its previously acquired knowledge base. This is evident in Table~\ref{tab:english_evaluation_without}, where its performance on English benchmarks drastically declines compared to the original base model, \textit{TinyLlama}.

In contrast, our proposed approach, denoted as \textit{Kuwain}, applies layer extension and selective training (training only newly added layers), aiming to integrate Arabic while preserving the model's prior capabilities. As shown in Table \ref{tab:english_evaluation_without} \textit{Kuwain }maintains strong performance on English benchmarks, while also achieving Arabic performance comparable to  \textit{Kuwain-Naive} as demonstrated in \ref{tab:arabic_evaluation_without}

This contrast highlights the strength of our strategy: it enables the model to expand into a new language (Arabic) without catastrophic forgetting of previously learned knowledge (English), thereby supporting more robust multilingual development.

\begin{table}[H]
\centering
\begin{minipage}[t]{0.48\textwidth}
\centering
\fontsize{10}{12}\selectfont
\captionsetup{justification=centering}
\begin{tabular}{lcc}
\hline
\textbf{Dataset} & \textbf{TinyLlama} & \textbf{Kuwain-Naive} \\
\hline
HellaSwag & 59.20 & 45.35 \\
Obqa & 36.00 & 29.20 \\
ARC-c & 30.10 & 25.68 \\
ARC-e & 55.25 & 45.24 \\
boolq & 57.83 & 61.90 \\
piqa & 73.29 & 66.70 \\
WinoGrande & 59.12 & 53.91 \\
\textbf{Avg} & \textbf{52.99} & 46.85 \\
\hline
\end{tabular}
\caption{English evaluation of \textit{TinyLlama} vs. \textit{Kuwain-Naive} (standard pre-training).}
\label{tab:english_evaluation_without}
\end{minipage}
\hfill
\begin{minipage}[t]{0.48\textwidth}
\centering
\fontsize{10}{12}\selectfont
\captionsetup{justification=centering}
\begin{tabular}{lcc}
\hline
\textbf{Dataset} & \textbf{Kuwain-Naive} & \textbf{Kuwain} \\
\hline
HellaSwag & 32.59 & 33.20 \\
Obqa & 27.80 & 27.60 \\
ARC-c & 25.26 & 25.49 \\
ARC-e & 36.44 & 36.76 \\
boolq & 61.80 & 62.35 \\
piqa & 54.06 & 54.35 \\
copa & 57.30 & 56.17 \\
\textbf{Avg} & \textbf{42.17} & 42.27 \\
\hline
\end{tabular}
\caption{Arabic evaluation of \textit{Kuwain-Naive} vs. our model \textit{Kuwain}.}
\label{tab:arabic_evaluation_without}
\end{minipage}
\end{table}

\section{Evaluation}
\label{sec:evaluation} 
This section presents a comprehensive evaluation of the \textit{Kuwain} model, highlighting its effectiveness in Arabic language tasks while maintaining robust English capabilities.

Table \ref{tab:arabic_evaluation} shows that \textit{Kuwain} significantly outperforms the base \textit{TinyLlama} \cite{zhang2024tinyllama} in Arabic language understanding and generation. These gains reflect the model's successful adaptation to Arabic with limited supervision, demonstrating the strength of our multilingual training approach.

\begin{table}[htb]
\centering
\small
\begin{tabular}{lcc}
\hline
\textbf{Dataset} & \textbf{TinyLlama} & \textbf{Kuwain} \\
\hline
HellaSwag & 29.90 & \textbf{37.14} \\
Obqa & 28.12 & \textbf{29.20} \\
ARC-c & 23.00 & \textbf{28.15} \\
ARC-e & 26.76 & \textbf{40.10} \\
boolq & 50.88 & \textbf{62.04} \\
piqa & 51.68 & \textbf{56.42} \\
copa & 48.31 & \textbf{58.38} \\
\textbf{Avg} & 36.95 & \textbf{44.49} \\
\midrule
\end{tabular}
\caption{\textbf{Arabic} Benchmark Evaluation of Kuwain and TinyLlama}
\label{tab:arabic_evaluation}
\end{table}

To contextualize these results, we evaluated \textit{Kuwain} on the Arabic language leaderboard \cite{OALL}, comparing it against a diverse set of multilingual and Arabic-focused models. As shown in Table \ref{tab:arabic_evaluation_with_all_model}, \textit{Kuwain} delivers competitive performance while being considerably smaller than many of its peers. This demonstrates the power of strategic data mixing and model tuning in low-resource scenarios, enabling high performance without inflating model size or training costs. Maintaining just 20\% of the original English data was sufficient to achieve strong Arabic performance while preserving English capabilities.

Our model, \textbf{Kuwain}, stands out as the smallest in Table \ref{tab:arabic_evaluation_with_all_model}, yet it demonstrates remarkably competitive performance when compared to models up to 10 times larger. This achievement is particularly noteworthy given the diverse nature of the compared models:
\begin{itemize}
    \item \textit{Arabic-centric Models}: Some models in the comparison, such as Jais \cite{sengupta2023jais}, AceGPT \cite{huang2023acegpt}, and ArabianGPT \cite{koubaa2024arabiangpt}, were specifically designed with a focus on Arabic language processing. These models underwent extensive pre-training and fine-tuning processes tailored for Arabic.
    \item \textit{Multilingual Models}: Other models in the comparison consider Arabic as one of many languages in their multilingual corpus, benefiting from broader language exposure during training.
\end{itemize}
Despite these differences, \textit{Kuwain} achieves comparable performance to both specialized and multilingual models. This is especially impressive considering our model's significantly smaller size, limited training data, and reduced computational requirements. Our approach demonstrates the potential for efficient, targeted language model expansion without the need for extensive resources or complete retraining.

\begin{table}[H]
\fontsize{10}{12}\selectfont
\begin{tabular}{lrrrrrrrrr}
\toprule
                              Model & Size & ARC-c &  ARC-e &  Boolq &  Copa &  HellaSwag &  Obqa &  Piqa &       Avg \\
\midrule
        gemma & 2B & 27.67 &  27.66 &  52.76 & 46.67 &      25.61 & 34.75 & 49.37 & 37.784286 \\
         falcon & 11B & 28.10 &  25.80 &  51.81 & 46.67 &      25.40 & 37.17 & 49.65 & 37.800000 \\
          falcon& 7B &  27.93 &  25.34 &  57.52 & 43.33 &      25.20 & 36.16 & 50.68 & 38.022857 \\
    ArabianGPT& 1.5B &  25.86 &  27.41 &  62.12 & 47.78 &      24.35 & 30.51 & 48.83 & 38.122857 \\
         falcon& 40B &  26.55 &  25.76 &  52.85 & 50.00 &      25.37 & 36.77 & 50.08 & 38.197143 \\
          bloom-1& 7B &  28.62 &  25.85 &  62.12 & 44.44 &      25.31 & 35.56 & 50.95 & 38.978571 \\
           jais& 13B &  28.53 &  28.43 &  62.12 & 48.89 &      25.67 & 35.35 & 54.56 & 40.507143 \\
         Qwen1.5& 4B &  29.31 &  28.09 &  62.76 & 52.22 &      25.07 & 36.16 & 50.85 & 40.637143 \\
          AceGPT& 7B &  29.66 &  28.64 &  62.36 & 48.89 &      25.89 & 38.38 & 52.59 & 40.915714 \\
    jais-adapted& 7B &  30.60 &  31.01 &  63.50 & 48.89 &      25.55 & 38.38 & 52.97 & 41.557143 \\
         Qwen1.5& 7B &  33.71 &  33.33 &  62.12 & 47.78 &      25.70 & 38.59 & 53.79 & 42.145714 \\
        jais-v1& 30B &  32.24 &  32.83 &  62.70 & 48.89 &      25.82 & 39.60 & 56.57 & 42.664286 \\
         AceGPT & 13B &  33.36 &  33.76 &  63.74 & 51.11 &      25.09 & 39.19 & 54.17 & 42.917143 \\
   jais-adapted& 13B &  33.62 &  34.90 &  65.03 & 47.78 &      26.41 & 39.39 & 54.99 & 43.160000 \\
    AceGPT-v1.5& 13B &  36.55 &  37.61 &  62.24 & 45.56 &      26.59 & 41.82 & 54.99 & 43.622857 \\
  Meta-Llama-3.1& 8B &  36.21 &  37.77 &  63.34 & 50.00 &      26.45 & 40.61 & 51.99 & 43.767143 \\
        Qwen1.5& 14B &  35.60 &  37.23 &  65.09 & 47.78 &      26.79 & 39.60 & 54.39 & 43.782857 \\
    Meta-Llama-3 & 8B &  35.69 &  39.00 &  62.12 & 50.00 &      26.65 & 42.63 & 53.68 & 44.252857 \\
    \hline
             \textbf{Kuwain}& $\downarrow$ 1.5B &  28.15 &  40.10 &  62.04 & 58.38 &      37.14 & 29.20 & 56.42 & 44.490000 \\
    \hline
        jais-v3 & 30B &  36.98 &  41.46 &  75.25 & 52.22 &      26.66 & 42.02 & 56.63 & 47.317143 \\
           gemma & 7B &  42.16 &  45.60 &  67.94 & 47.78 &      26.90 & 46.46 & 58.16 & 47.857143 \\
        Qwen1.5 & 32B &  43.19 &  43.99 &  64.08 & 51.11 &      29.19 & 46.87 & 59.36 & 48.255714 \\
          aya-23 & 8B &  40.09 &  40.31 &  77.91 & 54.44 &      30.86 & 44.44 & 64.70 & 50.392857 \\
   jais-adapted & 70B &  47.67 &  51.73 &  74.14 & 44.44 &      28.44 & 47.47 & 63.07 & 50.994286 \\
        Qwen1.5 & 72B &  49.05 &  46.40 &  65.61 & 51.11 &      34.02 & 47.27 & 64.76 & 51.174286 \\
         gemma-2 & 9B &  53.28 &  55.80 &  64.33 & 51.11 &      29.11 & 51.52 & 60.23 & 52.197143 \\
       Qwen1.5 & 110B &  51.81 &  51.18 &  62.12 & 50.00 &      36.65 & 50.91 & 63.72 & 52.341429 \\
 Meta-Llama-3.1 & 70B &  52.93 &  52.28 &  62.15 & 56.67 &      35.38 & 50.10 & 63.72 & 53.318571 \\
   Meta-Llama-3 & 70B &  54.91 &  55.80 &  62.18 & 56.67 &      35.03 & 51.72 & 62.79 & 54.157143 \\
        gemma-2 & 27B &  56.55 &  59.09 &  65.58 & 55.56 &      28.43 & 52.93 & 62.96 & 54.442857 \\
         aya-23 & 35B &  49.22 &  50.80 &  82.94 & 55.56 &      35.81 & 49.49 & 73.87 & 56.812857 \\
          Qwen2 & 72B & 56.55 &  57.45 &  69.97 & 55.56 &      38.33 & 57.17 & 69.67 & 57.814286 \\
 c4ai-command-r-v01 &  35B & 54.66 &  60.24 &  82.70 & 52.22 &      33.42 & 54.14 & 69.99 & 58.195714 \\
c4ai-command-r-plus & 104B & 57.76 &  60.87 &  85.00 & 52.22 &      37.88 & 52.73 & 70.92 & 59.625714 \\
\midrule
\end{tabular}
\caption{Arabic Leader board evaluation}
\label{tab:arabic_evaluation_with_all_model}
\end{table}

\section{Conclusion}
\label{sec:conclusion}
This research has demonstrated a novel and effective approach to expanding the linguistic capabilities of large language models while preserving their existing knowledge base. Through a carefully designed methodology incorporating vocabulary expansion, layer extension, and selective training (training only newly added layers), we have addressed a significant challenge in the field of multilingual natural language processing.

Key findings of our study include:
Preservation of Prior Knowledge: Our approach successfully maintained the model's performance on previously learned tasks, as evidenced by the English language evaluations. This stands in stark contrast to conventional training methods, which led to a substantial loss of prior knowledge.
Effective Acquisition of New Language: The model achieved comparable performance on the newly introduced language (Arabic), whether using our approach or conventional methods. This indicates that our method does not compromise the ability to learn new linguistic structures and vocabulary.
Based on the findings of our current research, we have identified several promising directions for future work:
\begin{itemize}
    \item \textit{Large-scale Arabic Data Collection:} We are actively working to collect and process a substantial amount of data, primarily in Arabic. This effort involves gathering information from various sources to create a rich and comprehensive dataset. This expanded data set will allow us to further test and refine our approach in a large-scale real-world scenario.
    \item \textit{Scaling Up the Approach:} We plan to extend our method to larger models. We hypothesize that performance in the newly integrated language is correlated with the original performance of the base model. By scaling up, we aim to test this hypothesis and further enhance the effectiveness of our approach across different model sizes and architectures.
\end{itemize}

These future directions aim to build on our current findings, potentially leading to more robust, versatile, and efficient multilingual language models.

\bibliographystyle{unsrt}
\bibliography{main}

\begin{thebibliography}{10}

\bibitem{touvron2023llama}
Hugo Touvron, Louis Martin, Kevin Stone, Peter Albert, Amjad Almahairi, Yasmine Babaei, Nikolay Bashlykov, Soumya Batra, Prajjwal Bhargava, Shruti Bhosale, et~al.
\newblock Llama 2: Open foundation and fine-tuned chat models.
\newblock {\em arXiv preprint arXiv:2307.09288}, 2023.

\bibitem{zhang2023don}
Xiang Zhang, Senyu Li, Bradley Hauer, Ning Shi, and Grzegorz Kondrak.
\newblock Don't trust chatgpt when your question is not in english: A study of multilingual abilities and types of llms.
\newblock {\em arXiv preprint arXiv:2305.16339}, 2023.

\bibitem{lai2023chatgpt}
Viet~Dac Lai, Nghia~Trung Ngo, Amir Pouran~Ben Veyseh, Hieu Man, Franck Dernoncourt, Trung Bui, and Thien~Huu Nguyen.
\newblock Chatgpt beyond english: Towards a comprehensive evaluation of large language models in multilingual learning.
\newblock {\em arXiv preprint arXiv:2304.05613}, 2023.

\bibitem{jiang2023mistral}
Albert~Q Jiang, Alexandre Sablayrolles, Arthur Mensch, Chris Bamford, Devendra~Singh Chaplot, Diego de~las Casas, Florian Bressand, Gianna Lengyel, Guillaume Lample, Lucile Saulnier, et~al.
\newblock Mistral 7b.
\newblock {\em arXiv preprint arXiv:2310.06825}, 2023.

\bibitem{abdin2024phi}
Marah Abdin, Sam~Ade Jacobs, Ammar~Ahmad Awan, Jyoti Aneja, Ahmed Awadallah, Hany Awadalla, Nguyen Bach, Amit Bahree, Arash Bakhtiari, Harkirat Behl, et~al.
\newblock Phi-3 technical report: A highly capable language model locally on your phone.
\newblock {\em arXiv preprint arXiv:2404.14219}, 2024.

\bibitem{team2024gemma}
Gemma Team, Thomas Mesnard, Cassidy Hardin, Robert Dadashi, Surya Bhupatiraju, Shreya Pathak, Laurent Sifre, Morgane Rivi{\`e}re, Mihir~Sanjay Kale, Juliette Love, et~al.
\newblock Gemma: Open models based on gemini research and technology.
\newblock {\em arXiv preprint arXiv:2403.08295}, 2024.

\bibitem{llama3}
Aaron Grattafiori, Abhimanyu Dubey, Abhinav Jauhri, Abhinav Pandey, Abhishek Kadian, Ahmad Al-Dahle, Aiesha Letman, Akhil Mathur, Alan Schelten, Alex Vaughan, et~al.
\newblock The llama 3 herd of models.
\newblock {\em arXiv preprint arXiv:2407.21783}, 2024.

\bibitem{zhang2024tinyllama}
Peiyuan Zhang, Guangtao Zeng, Tianduo Wang, and Wei Lu.
\newblock Tinyllama: An open-source small language model.
\newblock {\em arXiv preprint arXiv:2401.02385}, 2024.

\bibitem{ali2310tokenizer}
M~Ali, M~Fromm, K~Thellmann, R~Rutmann, M~L{\"u}bbering, J~Leveling, K~Klug, J~Ebert, N~Doll, JS~Buschhoff, et~al.
\newblock Tokenizer choice for llm training: Negligible or crucial? preprint (2023).
\newblock {\em arXiv preprint arXiv:2310.08754}, 2023.

\bibitem{hamed2025lahjawi}
Mohamed~Motasim Hamed, Muhammad Hreden, Khalil Hennara, Zeina Aldallal, Sara Chrouf, and Safwan AlModhayan.
\newblock Lahjawi: Arabic cross-dialect translator.
\newblock In {\em Proceedings of the 4th Workshop on Arabic Corpus Linguistics (WACL-4)}, pages 12--24, 2025.

\bibitem{chowdhery2023palm}
Aakanksha Chowdhery, Sharan Narang, Jacob Devlin, Maarten Bosma, Gaurav Mishra, Adam Roberts, Paul Barham, Hyung~Won Chung, Charles Sutton, Sebastian Gehrmann, et~al.
\newblock Palm: Scaling language modeling with pathways.
\newblock {\em Journal of Machine Learning Research}, 24(240):1--113, 2023.

\bibitem{le2023bloom}
Teven Le~Scao, Angela Fan, Christopher Akiki, Ellie Pavlick, Suzana Ili{\'c}, Daniel Hesslow, Roman Castagn{\'e}, Alexandra~Sasha Luccioni, Fran{\c{c}}ois Yvon, Matthias Gall{\'e}, et~al.
\newblock Bloom: A 176b-parameter open-access multilingual language model.
\newblock {\em https://inria.hal.science/hal-03850124}, 2023.

\bibitem{xue2020mt5}
L~Xue.
\newblock mt5: A massively multilingual pre-trained text-to-text transformer.
\newblock {\em arXiv preprint arXiv:2010.11934}, 2020.

\bibitem{blevins2022language}
Terra Blevins and Luke Zettlemoyer.
\newblock Language contamination helps explain the cross-lingual capabilities of english pretrained models.
\newblock {\em arXiv preprint arXiv:2204.08110}, 2022.

\bibitem{kaplan2020scaling}
Jared Kaplan, Sam McCandlish, Tom Henighan, Tom~B Brown, Benjamin Chess, Rewon Child, Scott Gray, Alec Radford, Jeffrey Wu, and Dario Amodei.
\newblock Scaling laws for neural language models.
\newblock {\em arXiv preprint arXiv:2001.08361}, 2020.

\bibitem{brown2020language}
Tom~B Brown.
\newblock Language models are few-shot learners.
\newblock {\em arXiv preprint arXiv:2005.14165}, 2020.

\bibitem{shaham2024multilingual}
Uri Shaham, Jonathan Herzig, Roee Aharoni, Idan Szpektor, Reut Tsarfaty, and Matan Eyal.
\newblock Multilingual instruction tuning with just a pinch of multilinguality.
\newblock {\em arXiv preprint arXiv:2401.01854}, 2024.

\bibitem{gupta2023continual}
Kshitij Gupta, Benjamin Th{\'e}rien, Adam Ibrahim, Mats~L Richter, Quentin Anthony, Eugene Belilovsky, Irina Rish, and Timoth{\'e}e Lesort.
\newblock Continual pre-training of large language models: How to (re) warm your model?
\newblock {\em arXiv preprint arXiv:2308.04014}, 2023.

\bibitem{cui2023efficient}
Yiming Cui, Ziqing Yang, and Xin Yao.
\newblock Efficient and effective text encoding for chinese llama and alpaca.
\newblock {\em arXiv preprint arXiv:2304.08177}, 2023.

\bibitem{pluster2023leolm}
Bj{\"o}rn Pl{\"u}ster.
\newblock Leolm: Igniting german-language llm research, 2023.

\bibitem{okazaki2024building}
Naoaki Okazaki, Kakeru Hattori, Hirai Shota, Hiroki Iida, Masanari Ohi, Kazuki Fujii, Taishi Nakamura, Mengsay Loem, Rio Yokota, and Sakae Mizuki.
\newblock Building a large japanese web corpus for large language models.
\newblock {\em arXiv preprint arXiv:2404.17733}, 2024.

\bibitem{voukoutis2024meltemi}
Leon Voukoutis, Dimitris Roussis, Georgios Paraskevopoulos, Sokratis Sofianopoulos, Prokopis Prokopidis, Vassilis Papavasileiou, Athanasios Katsamanis, Stelios Piperidis, and Vassilis Katsouros.
\newblock Meltemi: The first open large language model for greek.
\newblock {\em arXiv preprint arXiv:2407.20743}, 2024.

\bibitem{blevins2024breaking}
Terra Blevins, Tomasz Limisiewicz, Suchin Gururangan, Margaret Li, Hila Gonen, Noah~A Smith, and Luke Zettlemoyer.
\newblock Breaking the curse of multilinguality with cross-lingual expert language models.
\newblock {\em arXiv preprint arXiv:2401.10440}, 2024.

\bibitem{zhou2024moe}
Hao Zhou, Zhijun Wang, Shujian Huang, Xin Huang, Xue Han, Junlan Feng, Chao Deng, Weihua Luo, and Jiajun Chen.
\newblock Moe-lpr: Multilingual extension of large language models through mixture-of-experts with language priors routing.
\newblock {\em arXiv preprint arXiv:2408.11396}, 2024.

\bibitem{yong2022bloom+}
Zheng-Xin Yong, Hailey Schoelkopf, Niklas Muennighoff, Alham~Fikri Aji, David~Ifeoluwa Adelani, Khalid Almubarak, M~Saiful Bari, Lintang Sutawika, Jungo Kasai, Ahmed Baruwa, et~al.
\newblock Bloom+ 1: Adding language support to bloom for zero-shot prompting.
\newblock {\em arXiv preprint arXiv:2212.09535}, 2022.

\bibitem{sengupta2023jais}
Neha Sengupta, Sunil~Kumar Sahu, Bokang Jia, Satheesh Katipomu, Haonan Li, Fajri Koto, Osama~Mohammed Afzal, Samta Kamboj, Onkar Pandit, Rahul Pal, et~al.
\newblock Jais and jais-chat: Arabic-centric foundation and instruction-tuned open generative large language models.
\newblock {\em arXiv preprint arXiv:2308.16149}, 2023.

\bibitem{huang2023acegpt}
Huang Huang, Fei Yu, Jianqing Zhu, Xuening Sun, Hao Cheng, Dingjie Song, Zhihong Chen, Abdulmohsen Alharthi, Bang An, Ziche Liu, et~al.
\newblock Acegpt, localizing large language models in arabic.
\newblock {\em arXiv preprint arXiv:2309.12053}, 2023.

\bibitem{koubaa2024arabiangpt}
Anis Koubaa, Adel Ammar, Lahouari Ghouti, Omar Najar, and Serry Sibaee.
\newblock Arabiangpt: Native arabic gpt-based large language.
\newblock {\em arXiv preprint arXiv:2402.15313}, 2024.

\bibitem{saiful2024allam}
M~Saiful~Bari, Yazeed Alnumay, Norah~A Alzahrani, Nouf~M Alotaibi, Hisham~A Alyahya, Sultan AlRashed, Faisal~A Mirza, Shaykhah~Z Alsubaie, Hassan~A Alahmed, Ghadah Alabduljabbar, et~al.
\newblock Allam: Large language models for arabic and english.
\newblock {\em arXiv e-prints}, pages arXiv--2407, 2024.

\bibitem{culturax}
Thuat Nguyen, Chien Van~Nguyen, Viet~Dac Lai, Hieu Man, Nghia~Trung Ngo, Franck Dernoncourt, Ryan~A Rossi, and Thien~Huu Nguyen.
\newblock Culturax: A cleaned, enormous, and multilingual dataset for large language models in 167 languages.
\newblock {\em arXiv preprint arXiv:2309.09400}, 2023.

\bibitem{c4}
Colin Raffel, Noam Shazeer, Adam Roberts, Katherine Lee, Sharan Narang, Michael Matena, Yanqi Zhou, Wei Li, and Peter~J. Liu.
\newblock Exploring the limits of transfer learning with a unified text-to-text transformer.
\newblock {\em arXiv e-prints}, 2019.

\bibitem{wu2024llama}
Chengyue Wu, Yukang Gan, Yixiao Ge, Zeyu Lu, Jiahao Wang, Ye~Feng, Ping Luo, and Ying Shan.
\newblock Llama pro: Progressive llama with block expansion.
\newblock {\em arXiv preprint arXiv:2401.02415}, 2024.

\bibitem{shazeer2020glu}
Noam Shazeer.
\newblock Glu variants improve transformer.
\newblock {\em arXiv preprint arXiv:2002.05202}, 2020.

\bibitem{csaki2023efficiently}
Zoltan Csaki, Pian Pawakapan, Urmish Thakker, and Qiantong Xu.
\newblock Efficiently adapting pretrained language models to new languages.
\newblock {\em arXiv preprint arXiv:2311.05741}, 2023.

\bibitem{nikolich2024vikhr}
Aleksandr Nikolich, Konstantin Korolev, and Artem Shelmanov.
\newblock Vikhr: The family of open-source instruction-tuned large language models for russian.
\newblock {\em arXiv preprint arXiv:2405.13929}, 2024.

\bibitem{farghaly2009arabic}
Ali Farghaly and Khaled Shaalan.
\newblock Arabic natural language processing: Challenges and solutions.
\newblock {\em ACM Transactions on Asian Language Information Processing (TALIP)}, 8(4):1--22, 2009.

\bibitem{elkateb2006arabic}
Sabri Elkateb, William~J Black, Piek Vossen, David Farwell, Horacio Rodriguez, Adam Pease, Musa Alkhalifa, and Christiane Fellbaum.
\newblock Arabic wordnet and the challenges of arabic.
\newblock In {\em Proceedings of the International Conference on the Challenge of Arabic for NLP/MT}, pages 15--24, 2006.

\bibitem{issa2023morphological}
Iyad Issa.
\newblock Morphological complexity in arabic spelling and its implication for cognitive processing.
\newblock {\em Journal of Psycholinguistic Research}, 52(1):331--357, 2023.

\bibitem{kudo1808sentencepiece}
Taku Kudo and John Richardson.
\newblock Sentencepiece: A simple and language independent subword tokenizer and detokenizer for neural text processing. arxiv 2018.
\newblock {\em arXiv preprint arXiv:1808.06226}, 1808.

\bibitem{gromov2024unreasonable}
Andrey Gromov, Kushal Tirumala, Hassan Shapourian, Paolo Glorioso, and Daniel~A Roberts.
\newblock The unreasonable ineffectiveness of the deeper layers.
\newblock {\em arXiv preprint arXiv:2403.17887}, 2024.

\bibitem{antoun2020arabert}
Wissam Antoun, Fady Baly, and Hazem Hajj.
\newblock Arabert: Transformer-based model for arabic language understanding.
\newblock {\em arXiv preprint arXiv:2003.00104}, 2020.

\bibitem{tao2024scaling}
Chaofan Tao, Qian Liu, Longxu Dou, Niklas Muennighoff, Zhongwei Wan, Ping Luo, Min Lin, and Ngai Wong.
\newblock Scaling laws with vocabulary: Larger models deserve larger vocabularies.
\newblock {\em arXiv preprint arXiv:2407.13623}, 2024.

\bibitem{bostrom2020byte}
Kaj Bostrom and Greg Durrett.
\newblock Byte pair encoding is suboptimal for language model pretraining.
\newblock {\em arXiv preprint arXiv:2004.03720}, 2020.

\bibitem{OALL}
Ali Elfilali, Hamza Alobeidli, Clémentine Fourrier, Basma El~Amel Boussaha, Ruxandra Cojocaru, Nathan Habib, and Hakim Hacid.
\newblock Open arabic llm leaderboard.
\newblock \url{https://huggingface.co/spaces/OALL/Open-Arabic-LLM-Leaderboard}, 2024.

\end{thebibliography}

\clearpage
\newpage
\appendix

\setcounter{table}{0}
\setcounter{figure}{0}
\renewcommand{\thetable}{A\arabic{table}}
\renewcommand{\thefigure}{A\arabic{figure}}

\end{document}